\documentclass[10pt,twocolumn,letterpaper]{article}

\usepackage{cvpr}      

\usepackage{subfiles}
\usepackage{graphicx}
\usepackage{amsmath}
\usepackage{amssymb}
\usepackage{booktabs}
\usepackage{multirow}
\usepackage{bm}
\usepackage{color}
\usepackage{tabularx}
\usepackage{adjustbox}
\usepackage[accsupp]{axessibility}  

\let\vec\boldvec
\let\mat\boldvec
\newcolumntype{P}[1]{>{\centering\arraybackslash}p{#1}}

\definecolor{Light1}{rgb}{0.98, 0.95, 0.90}
\definecolor{Light2}{rgb}{0.98, 0.98, 0.93}
\definecolor{Light3}{rgb}{0.98, 0.98, 1}
\definecolor{Light4}{rgb}{0.98, 0.96, 0.95}
\usepackage[table]{xcolor}

\usepackage[pagebackref,breaklinks,colorlinks]{hyperref}

\usepackage[capitalize]{cleveref}
\crefname{section}{Sec.}{Secs.}
\Crefname{section}{Section}{Sections}
\Crefname{table}{Table}{Tables}
\crefname{table}{Tab.}{Tabs.}

\def\cvprPaperID{2195} 
\def\confName{CVPR}
\def\confYear{2022}

\begin{document}

\title{iPLAN: Interactive and Procedural Layout Planning}

\author{Feixiang He\\
University of Leeds, UK\\
{\tt\small scfh@leeds.ac.uk}
\and
Yanlong Huang\\
University of Leeds, UK\\
{\tt\small y.l.huang@leeds.ac.uk}

\and
He Wang \thanks{Corresponding author}\\
University of Leeds, UK\\
{\tt\small h.e.wang@leeds.ac.uk}
}

\maketitle

\begin{abstract}
Layout design is ubiquitous in many applications, e.g. architecture/urban planning, etc, which involves a lengthy iterative design process. Recently, deep learning has been leveraged to automatically generate layouts via image generation, showing a huge potential to free designers from laborious routines.
While automatic generation can greatly boost productivity, designer input is undoubtedly crucial. An ideal AI-aided design tool should automate repetitive routines, and meanwhile
accept human guidance and provide smart/proactive suggestions. However, the capability of involving humans into the loop has been largely ignored in existing methods which are mostly end-to-end approaches. To this end, we propose a new human-in-the-loop generative model, \textbf{iPLAN}, which is capable of automatically generating layouts, but also interacting with designers throughout the whole procedure, enabling humans and AI to co-evolve a sketchy idea gradually into the final design. iPLAN is evaluated on diverse datasets and compared with existing methods. The results show that iPLAN has high \textbf{fidelity} in producing similar layouts to those from human designers, great \textbf{flexibility} in accepting designer inputs and providing design suggestions accordingly, and strong \textbf{generalizability} when facing unseen design tasks and limited training data.

\end{abstract}

\section{Introduction}

Layout generation has recently spiked research interests in computer vision/graphics, aiming to automate the design process and boost the productivity. The traditional design process follows a diagram of iteratively adjusting/finalizing details from coarse to fine and global to local, which imposes repetitive and laborious routines on designers. Very recently, it has been shown that automatic image generation of such designs (with minimal human input) is possible through learning from data~\cite{nauata2020house,nauata2021house,Wu_DeepLayout_2019,hu2020graph2plan}. This new line of research combines deep learning with design and has demonstrated a new avenue for AI-aided design.

To achieve full automation, current research tends to learn from existing designs in an end-to-end fashion, and then to generate new ones with qualitative similarity and sufficient diversity. Taking floorplan as an example, automated generation can be based on simple human input, such as the boundary of the floor space~\cite{Wu_DeepLayout_2019}, the relations among rooms~\cite{nauata2021house, nauata2020house}, or both~\cite{hu2020graph2plan}. While fully automated generation is important, design is in nature a procedural process, which involves alternations between repetitive routines and creative thinking at multiple intermediate stages~\cite{rengel2011interior}. Therefore, an ideal AI-aided system should automate the routine part while allowing the designer to impart creativity. This requires the system to be able to \textit{interact} with the designer, in the sense that it should accept designer's guidance, then actively suggest possible solutions accordingly, completing a feed-back loop. So far, the human-in-the-loop element is largely missing, which prevents a closer integration of AI and existing design practice.

Designing such an AI model faces several intrinsic challenges. In practice, learning how to interact with the designer requires a full observation of the decisions made at every intermediate stage. However, existing datasets, such as RPLAN~\cite{Wu_DeepLayout_2019} and LIFULL \cite{lifull}, usually only include the final designs, without the stage-to-stage design process. One potential solution to overcome this issue is to reverse-engineer intermediate stages from final designs, which however leads to another difficulty: the order of stages depends on the specific task/goal and could vary dramatically even for the same final design. Further, the order uncertainty is exacerbated by the strong personal styles and preferences of designers. Thus, how to design an AI system that can account for the above factors is a key research question, which is under-explored to date.


In this paper, we propose a new human-in-the-loop generative model for layout generation, which is referred to as \textit{interactive planning} (iPLAN).  Unlike previous work, iPLAN is equipped with a user-friendly interaction 
mechanism, which is achieved by letting the AI model learn the multi-stage design process, aiming to accommodate free-form \textit{user interactions} and propose \textit{design suggestions} at every stage. This allows designer inputs at different stages across a wide range of levels of detail,
while offering the capability of fully automated generation. To address the challenge of missing procedural design data, we reverse-engineer the final design to obtain the stage-to-stage process, based on principles that are widely adopted by professional designers~\cite{rengel2011interior}. This enables us to design a Markov chain model to capture the full design procedure. Since there is more than one way to reverse-engineer the final designs (\ie, the stage order can vary), our model is designed with the capacity of accepting inputs with an arbitrary order, and consequently can learn the style variations implicitly from the data.

While iPLAN is general, we focus on floorplan design in this paper. iPLAN has been validated on two large-scale benchmark datasets, \ie, RPLAN\cite{Wu_DeepLayout_2019} and LIFULL \cite{lifull}, under diverse scenarios. The experiments show that our model is \textit{versatile} in accepting designer inputs at various levels of detail, from minimal input and automatic generation, to stage-to-stage human guidance and interactive design. By learning from designs augmented by reverse-engineered processes, our model exhibits \textit{high fidelity} in generating new designs with close style similarity and sufficient diversity. Finally, our model is highly \textit{flexible} and \textit{generalizable} when trained on varying amounts of data and facing unseen spaces and design requirements that are categorically different from the training data.

\textbf{Contributions:}
{(\emph{i})} We propose a novel human-in-the-loop generative model iPLAN which respects design principles and mimics the design styles of professional designers implicitly. {(\emph{ii})} We demonstrate a successful fine-grained stage-to-stage generative model for floorplan, as opposed to existing end-to-end approaches. {(\emph{iii})} We show a variety of design scenarios, including fully automated generation, interactive planning with user instructions, and generalization for unseen tasks; {(\emph{iv})} We conduct extensive evaluations on diverse benchmark datasets and demonstrate that iPLAN outperforms the state of the art under multiple metrics.

\section{Related Work}
Layout generation has been an active research area in computer vision, \eg, indoor scene synthesis\cite{yu2011make,Zhao2014indexing,zhao2016relationship,merrell2011interactive,fisher2012example} and floorplan generation \cite{harada1995interactive, bao2013generating, muller2006procedural, peng2014computing, Wu_DeepLayout_2019, hu2020graph2plan}, image composition\cite{johnson2018image, ashual2019specifying}, \etc. Existing approaches can be generally grouped into two categories: handcrafted rule-based methods and data-driven methods. We mainly review the latter as they are closely related to our research.

\textit{Indoor scene synthesis}. The synthesis of indoor scenes typically
involves the placement of furniture models from an existing database into a given room. Convolutional neural networks can be trained to iteratively insert one object at a time into a room for indoor scene generation \cite{wang2018deep, ritchie2019fast}. High-level scene semantics can also be employed, \eg, scene graphs, as a prior for a more controlled generation~\cite{wang2019planit}. The biggest difference between indoor scene synthesis and floorplan generation is their requirements on the space partitioning. While indoor scene synthesis places objects in a room where the room itself does not need to be divided, floorplan normally requires explicit space division for different functionalities.

\textit{Image composition from scene graphs}. Another related field is image composition from scene graphs, where the task is to derive the scene from a layout graph that describes the locations and features of the objects. Such generation can be achieved by Generative Adversarial Networks (GANs) based on graph convolution~\cite{johnson2018image}. Further improvements can be obtained by separating the layout from the appearance of the objects~\cite{ashual2019specifying}. For more controllability, Li \etal~\cite{li2019pastegan} synthesize images from a scene graph and the corresponding image crops. In contrast to the floorplan, the challenge in image composition is how to compose different objects into an image rather than partitioning the space.

\textit{Floorplan generation}
Floorplan generation can be formulated as an image synthesis problem which is one active research area in computer vision. Due to the surge of deep learning, the most promising approaches are GANs \cite{goodfellow2014generative, karras2017progressive, zhang2019self, karras2020analyzing, karras2019style, brock2018large,karras2020training}. 
Image-based GANs have been proven effective in floorplan generation~\cite{chaillou2020archigan,isola2017image,zhu2017unpaired, zhu2017multimodal, richardson2021encoding,zhu2020sean,huang2018multimodal}.
Graph-based GANs can also produce floorplans by only taking spacial constraints, such as room connections, room types, in the form of graph~\cite{nauata2020house, nauata2021house}. However, all these methods are end-to-end approaches, and therefore provide limited interactivity to the designer.

More recently, some human-in-the-loop approaches are proposed.  Wu \etal~\cite{Wu_DeepLayout_2019} propose a two-phase approach to produce floorplans of residential buildings. The model successively predicts locations of rooms and walls given a building boundary, and converts the predicted layout into a vector graphics format. Graph2Plan \cite{hu2020graph2plan} combines the topology information in the form of graphs with spatial constraints, to instantiate rooms accordingly. These methods enable human interactions at certain stages, \eg, modifying room locations and retrieving the graph. Different from the existing methods, we propose a fine-grained generative model which enables interactions with the designer at different levels, from providing high-level design requirements to low-level instructions at
a step.

\section{Methodology}
\begin{figure}[tb]
  \centering
  \includegraphics[width=0.5\textwidth]{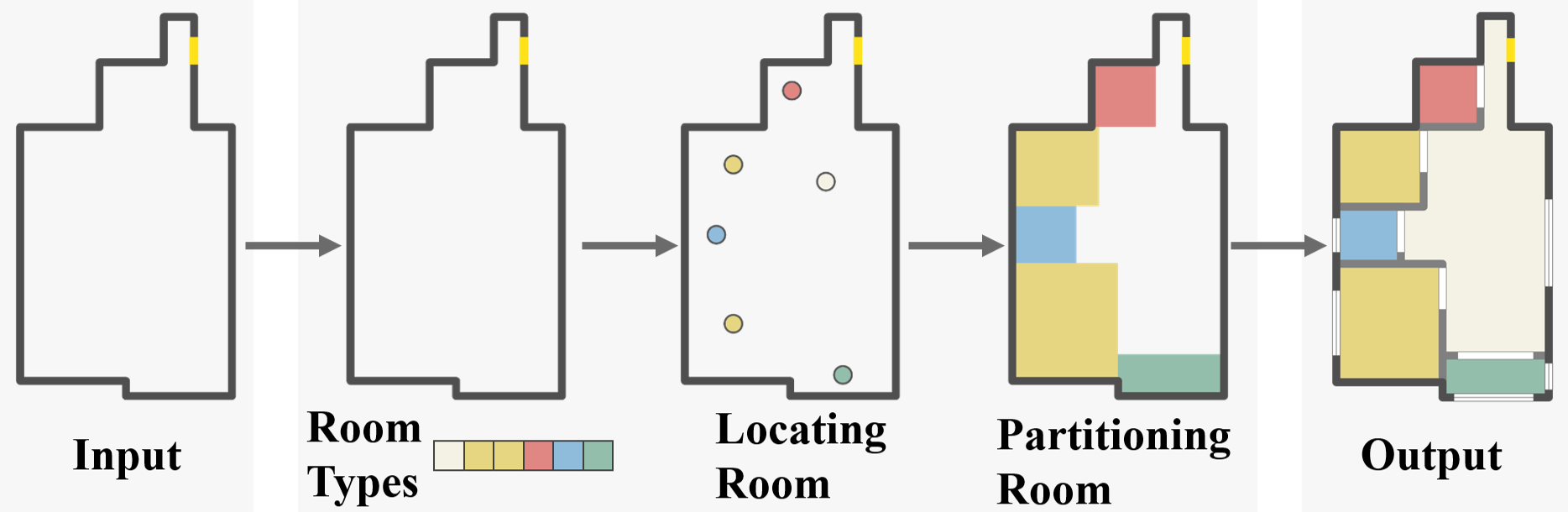}
   \caption{Overview of our framework iPLAN. Room types are predicted at a time, while room locations and partitions are predicted iteratively.}
   \label{fig:framework}
   \vspace{-0.4cm}
\end{figure}

The overview of iPLAN is presented in Fig.~\ref{fig:framework}. Without loss of generality, our model takes the space boundary as input, and decomposes the design procedure into: acquiring room types, locating rooms, and finalizing room partitions. Such workflow aims to mimic human designers~\cite{rengel2011interior} and 
accept designer inputs at any stage. The workflow is modeled  as a joint probability distribution of all the aforementioned factors, which is then factorized into stages and later formulated as a Markov chain (\cref{sec:problem}). Next, each factorized distribution provides a flexible entry point to incorporate user input or can be used for automatic generation (\cref{sec:housetype}--\cref{sec:roomPartitioning}).

\subsection{Problem Formulation}
\label{sec:problem}

A dataset of $H$ layouts is denoted by $\mat{D}= \{\vec{D}_{i}\}_{i=1}^{H}$ with the $i$-th layout $\vec{D}_{i}=(\mat{B}_{i},\mat{R}_{i}, \vec{T}_{i}, N_{i}, \mat{C}_{i})$. $\mat{B}_{i} \in \mathbb{R}^{128\times128}$ is the boundary,
$N_i$ denotes the total number of rooms. We use $j$ to index a specific room. 
$\mat{R}_{i}=\{\vec{r}_{i,j}\}_{j=1}^{N_i}$
denotes the room regions with 
$\vec{r}_{i,j} \in \mathbb{R}^4$
indicating the top-left and bottom-right corners of the bounding box of the room. $\vec{T}_{i}=\{t_{i,j}\}_{j=1}^{N_i}$
is a set of room types $t_{i,j} \in \mathbb{Z}^{+}$. $\mat{C}_{i}=\{\vec{c}_{i,j}\}_{j=1}^{N_i}$ is a set of room centers $\vec{c}_{i,j}\in \mathbb{R}^{2}$. Given $\mat{D}$, we aim to design a generative model for $\mathcal{P}(\mat{D})=\prod_{i=1}^H\mathcal{P}(\vec{D}_i)$. 

Similarly to existing methods, we also formulate floorplan design as an image generation problem, but with our focus on 
proposing a fine-grained generative model to enable human-in-the-loop interaction, through decomposing $\mathcal{P}(\mat{D})$ appropriately. From a mathematical perspective, there are many ways to decompose $\mathcal{P}(\mat{D})$. 
In order to allow human inputs across various levels of detail, we rely on design principles and widely adopted practice~\cite{rengel2011interior} to mimic the human design workflow, and naturally divide a floorplan design procedure into several stages.
First, the desired number of rooms and their types are determined. Next, locations of areas with specific functionality (\eg, living rooms, bedrooms) are roughly estimated. Finally, room partitioning is conducted to finalize the design. This design diagram serves as a strong inductive bias in our model, following which $\mathcal{P}(\mat{D})$ is decomposed into:
\begin{equation}
\begin{aligned}  
\mathcal{P}(\vec{D})=\prod_{i=1}^H \mathcal{P}(\vec{D}_i)
=\prod_{i=1}^H\mathcal{P}(\mat{R}_i,\mat{C}_i,\vec{T}_i,N_i,\mat{B}_i) \\
=\prod_{i=1}^H\mathcal{P}(\mat{R}_i|\mat{C}_i,\vec{T}_i,N_i,\mat{B}_i)\mathcal{P}(\mat{C}_i|\vec{T}_i,N_i,\mat{B}_i) \\
\mathcal{P}(\vec{T}_i,N_i|\mat{B}_i)\mathcal{P}(\mat{B}_i)
\end{aligned}
\label{eq:joint_distribution}
\end{equation}
where $\mathcal{P}(\mat{B}_i)$ accounts for the boundary known \textit{a priori}; $\mathcal{P}(\vec{T}_i,N_i|\mat{B}_i)$ is to infer the desired number of rooms and room types; $\mathcal{P}(\mat{C}_i|\vec{T}_i,N_i,\mat{B}_i)$ and $\mathcal{P}(\mat{R}_i|\mat{C}_i,\vec{T}_i,N_i,\mat{B}_i)$ 
respectively correspond to the coarse and fine designs of the layout, where the former estimates the room locations while the latter predicts the exact partitions. A visualization of \cref{eq:joint_distribution} is provided in~\Cref{fig:framework},
where the second, third and fourth blocks correspond to $\mathcal{P}(\vec{T}_i,N_i|\mat{B}_i)$, $\mathcal{P}(\mat{C}_i|\vec{T}_i,N_i,\mat{B}_i)$ and $\mathcal{P}(\mat{R}_i|\mat{C}_i,\vec{T}_i,N_i,\mat{B}_i)$, respectively.

Inspired by indoor scene synthesis where objects are placed iteratively\cite{wang2018deep, ritchie2019fast}, we assume rooms are designed one by one. Formally, we model ${P}(\mat{R}_i|\mat{C}_i,\vec{T}_i,N_i,\mat{B}_i)$ and $\mathcal{P}(\mat{C}_i|\vec{T}_i,N_i,\mat{B}_i)$ as Markov chains, \ie, designs are conducted in a step-wise manner and early decisions will affect later ones, which allows a designer to focus on one room at a time and give guidance at any step:
\begin{align}
&\mathcal{P}(\mat{C}_i|\vec{T}_i,N_i,\mat{B}_i)= \prod_{j=1}^{N_i} \mathcal{P}(\vec{c}_{i,j}|\vec{c}_{i,<j },t_{i,j}, N_i,\mat{B}_i)
\label{eq:roomcenter_distribution} \\
&\mathcal{P}\!(\mat{R}_i |\mat{C}_i,\vec{T}_i, N_i, \mat{B}_i)
\!=\!\prod_{j=1}^{N_i}\! \mathcal{P}\!(\vec{r}_{i,j}|\vec{r}_{i,<j}, \vec{c}_{i,j},t_{i,j}, N_i, \mat{B}_i)
\label{eq:room_area_distribution}
\end{align}
where $\vec{r}_{i,<j}\!\!=\!\!\{\!\vec{r}_{i,1},\!\ldots,\! \vec{r}_{i,j-1}\!\}$ and $\vec{c}_{i,<j} \!\!=\!\! \{\!\vec{c}_{i,1},\ldots, \vec{c}_{i,j-1}\!\}$ denote the set of allocated room partitioning and centers before the $j$-th room, respectively.

\textbf{Connections to existing research}. Eq.~(\ref{eq:joint_distribution})--(\ref{eq:room_area_distribution}) is a generalization of existing methods and is more fine-grained. Graph2Plan\cite{hu2020graph2plan} simultaneously determines $\mat{C}_i$, $\mat{T}_i$ and $N_i$ given $\mat{B}_i$, and then $\mathcal{P}(\mat{R}_i  |\mat{C}_i,\vec{T}_i, N_i, \mat{B}_i)$ is predicted. 
RPLAN\cite{Wu_DeepLayout_2019} 
estimates
$\mathcal{P}(\mat{C}_i,\mat{T}_i, N_i|\mat{B}_i)$ via predicting $\mathcal{P}(\vec{c}_{i,j},{t}_{i,j}|\vec{c}_{i,<j},{t}_{i,<j},\mat{B}_i)$ consecutively, and then indirectly estimates room areas $\mat{R}_i$ by locating walls. In contrast, we further decompose $\mathcal{P}(\vec{C}_i, \vec{T}_i,  N_i | \mat{B}_i)$ into $\mathcal{P}(\vec{T}_i,N_i|\mat{B}_i)$ and $\mathcal{P}(\mat{C}_i|\vec{T},N_i,\mat{B}_i)$, and further decompose the latter using \cref{eq:roomcenter_distribution}. Moreover, $\mathcal{P}(\mat{R}_i |\mat{C}_i,\vec{T}_i, N_i, \mat{B}_i)$ is also decomposed into multiple steps by \cref{eq:room_area_distribution}. Our decompositions bring a more fine-grained procedural generative model that allows for user interactions at arbitrary steps. This enables more flexible and closer human-AI interactions. For the sake of brevity, we omit the subscripts $i$ of $\mat{T}_i,N_i,\mat{B}_i$ and $\mat{C}_i$
in the following sections.

\subsection{The Number and Types of Rooms}
\label{sec:housetype}

$\bm{T}$ and $N$ are normally given beforehand. However, given a specific $\mat{B}$, the design might not be unique, \eg, the same space can be designed as a 2-bed or 3-bed flat. In other words, there exists a distribution of possible designs. Thus, we propose to learn their distributions $\mathcal{P}(\vec{T},N|\mat{B})$ from real designs by professional designers to enable automatic exploration. 
$\{\vec{T},N\}$ can be replaced by a random variable $\bm{Q}=\{q_{k}\}_{k=1}^{K}$, where $K$ denotes the number of room types in $\mat{D}$ and $q_{k}$ corresponds to the number of rooms under the $k$-th type. So, we model $P(\vec{Q}|\mat{B})$ instead.

We propose a boundary-conditioned Variational Autoencoder (BCVAE) based on VAE~\cite{sohn2015learning} where $\mat{B}$ serves as a condition. The model consists of an embedding module $\mathcal{F}_{ed}$, an encoder $\mathcal{F}_{en}$ and a decoder $\mathcal{F}_{de}$. By feeding $\mathcal{F}_{ed}$ with $\mat{B}$, it outputs an embedded vector $\vec{\gamma} \in \mathbb{R}^{128}$. Also, $\{\vec{\mu}, \vec{\Sigma}\} = \mathcal{F}_{en}(\vec{Q},\vec{\gamma})$ where $\vec{\mu} \in \mathbb{R}^{32}$ and $\vec{\Sigma} \in \mathbb{R}^{32\times32}$ are the mean and covariance of a Gaussian distribution. Furthermore, a latent variable $\vec{z}$ is sampled from $\mathcal{N}(\vec{\mu},\vec{\Sigma})$ using the reparameterization trick\cite{kingma2013auto}. Given $\vec{z}$ and $\vec{\gamma}$, we reconstruct $\vec{Q}$ by $\vec{\hat{Q}} = \mathcal{F}_{de}(\vec{z}, \vec{\gamma})$. 
We employ a standard VAE loss for training. The detailed network architecture and the training loss are discussed in the supplementary material.

During inference, for a new boundary $\vec{B}$, we sample $\vec{z}$ from $\mathcal{N}(\vec{0},\vec{I})$ and predict $\vec{\hat{Q}}$ by 
$\vec{\hat{Q}}= \mathcal{F}_{de}(\vec{z},\mathcal{F}_{ed}(\vec{B}))$, which is then used to recover $\vec{\hat{T}}$ and $\hat{N}$. 

\begin{figure}[tb]
  \centering
  \includegraphics[width=0.5\textwidth]{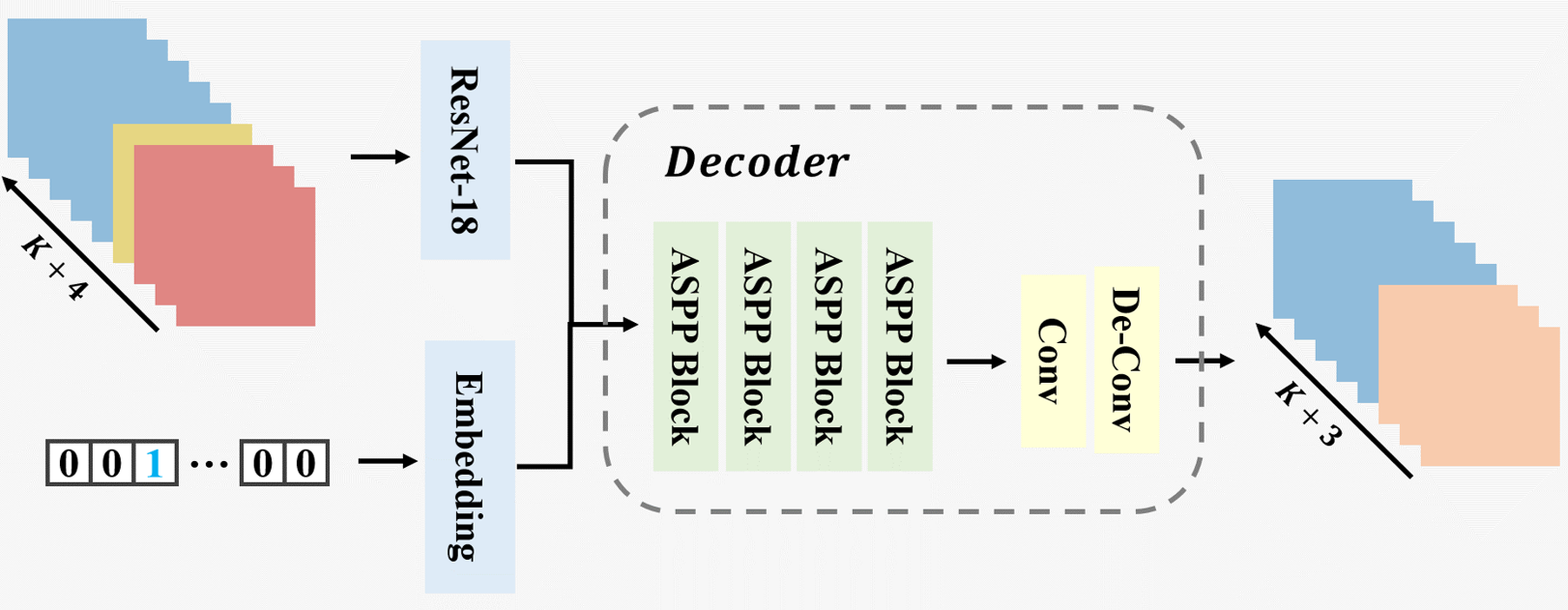}
   \caption{Prediction of a room center at one step}
   \label{fig:room_center_prediction}
   \vspace{-0.4cm}
\end{figure}

\subsection{Locating Rooms}
\label{sec:roomLocation}
Locating room regions $\mathcal{P}(\mat{C}|\vec{T},N,\mat{B})$ plays an essential role in the design process. Similarly to \cite{Wu_DeepLayout_2019}, we model room region prediction as a step-wise classification task. At each step, given a multi-channel image representation of the current design state and the next desired room type $t_{j}$, we predict the center of the next room. The multi-channel image encodes the boundary $\mat{B}$ and all previously predicted room centers $\vec{c}_{<j}$. The image consists of $K + 4$ binary channels, three of which label $\mat{B}$, \ie, the boundary, the front door and the interior area pixels. $K$ channels represent the predicted room centers, with each channel corresponding to a room type. For each room, its center is represented by a $9 \times 9$ square of pixels with value 1. We also use a channel to summarize the centers of all predicted rooms. Regarding the desired room type $t_{j}$, we convert it into a one-hot vector. 

We feed the multi-channel representation to a Resnet-18\cite{he2016deep} network and the one-hot vector of the desired room type to an embedding network to extract features. The embedding network has three fully-connected layers, followed by four convolution blocks, a Batch Normalization Layer and a LeakyReLu Layer. The extracted features are concatenated and subsequently fed to a decoder module that contains 4 atrous spatial pyramid pooling (ASPP) \cite{chen2017deeplab}, a Convolutional block and a Deconvolutional block. The output is $\mat{O} \in \mathbb{R}^{(K+3)\times 128 \times 128}$, giving a probability vector of all pixels. For each pixel, we predict $K+3$ labels, \ie, $K$ room types and EXISTING, FREE, and OUTSIDE, where EXISTING, FREE and OUTSIDE indicate whether a pixel belongs to an existing room, a free space (in the interior of the boundary) and the exterior, respectively. The whole model is shown in \Cref{fig:room_center_prediction}.

\label{sec:roomPartitioning}
\begin{figure}[tb]
  \centering
  \includegraphics[width=0.5\textwidth]{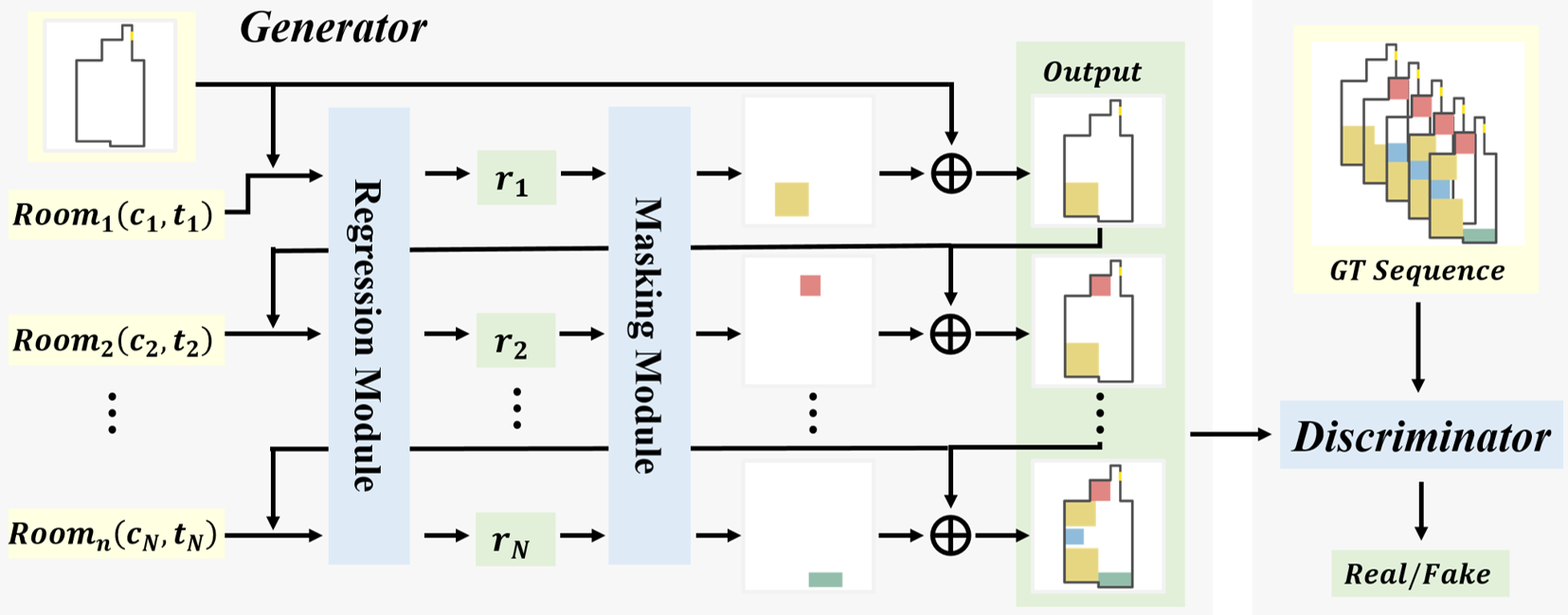}
   \caption{Procedural room shape generation}
   \label{fig:room_shape_generation}
   \vspace{-0.4cm}
\end{figure}

For training, we predict one room at a time, by decomposing the final design into a series of design states with one room added at each stage. However, there are multiple possible sequences for a given final design and the ground-truth step-to-step decisions are unavailable. Therefore, we employ a stochastic training process to learn all possible sequences by randomly removing rooms from a design. We propose a pixel-wise cross-entropy loss:
\begin{equation}
    \mathcal{L} = \sum_{h=1}^{128}\sum_{w=1}^{128} -\omega_{y}log\frac{exp({O}_{y,h,w})}{\sum_{k=1}^{K+3}exp({O}_{k,h,w})},
\end{equation}
where $y$ is the ground-truth class index for the pixel located at $(h, w)$.
$\omega_{y}$ represents the weight of $y$-th label, we set it to be 2 for $K$ room types and 1.25 for the other three labels. 

During the inference phase, for a sequence of room types ${\vec{T}}=\{{t}_1,{t}_2,\ldots,{t}_{{N}}\}$
we first predict the room center for ${t}_1$, \ie, $\mathcal{P}({\vec{c}}_{1}| {t}_{1}, {N},\mat{B})$, then we update the current design state using this predicted center $\hat{\vec{c}}_{1}$ and continue to predict the next room center for ${t}_2$, \ie, $\mathcal{P}({\vec{c}}_{2}| \hat{\vec{c}}_{1} ,{t}_{2}, {N},\mat{B})$. This procedure is repeated until all centers $\hat{\vec{C}}$ are determined. Note that when the order of elements in ${\vec{T}}$ varies, its corresponding layout can also change, thus providing design diversity. Also, at any intermediate step, user input (\eg, adjust a room center) can be incorporated into our model before it predicts the next room center.

\subsection{Predicting Room Partitioning}

After predicting the room centers $\hat{\vec{C}}$, we are ready for detailed room partitioning $\mathcal{P}({\mat{R}} |\hat{\mat{C}},\hat{\vec{T}}, \hat{\vec{N}}, {\mat{B}})$, where we need to consider the room size and the room shape. The room size is directly related to the specific functionality of the room, \eg, living rooms are normally larger than bathrooms for social interactions. The shape is affected by the functionality too but also strongly affected by the boundary geometry and the inner walls. Further, there is usually no `optimal' solution but a distribution of near-optimal solutions when looking at the final designs. Since the distribution can be arbitrary, we propose a new Generative Adversarial Network (GAN) to model the step-wise prediction of room partitioning. The general model is shown in~\Cref{fig:room_shape_generation}. Instead of directly predicting the shape of each room, we first predict its bounding box. This allows the bounding box predicted at a later stage to grab the space that is already allocated. We find this is a straightforward yet effective way to generate non-box rooms.

\textit{\textbf{Generator:}} The generator begins with a bounding-box regressor $\mathcal{F}_{b}$ that outputs the top-left and bottom-right corners of the box. $\mathcal{F}_{b}$ consists of six convolutional units and two fully connected layers, each unit with a Convolution layer, a Layer-Normalization layer and a ReLU layer. $\mathcal{F}_{b}$ enables the designer to interact with the system easily, \eg, relocating a room, modifying a room bounding box. However, only predicting the corners of bounding boxes is not convenient. We want to label all the pixels within the box so that we can work in the image space consistently. So we design a tailor-made masking module $\mathcal{F}_{m}$ to map each room bounding box into a room mask image.  $\mathcal{F}_{m}$ contains six convolution layers with $3 \times 3$ kernel, followed by a Sigmoid activation function. 

For a given sequence of room centers and types $\{({\vec{c}}_1, {t}_1), ({\vec{c}}_2, {t}_2), \ldots, ({\vec{c}}_{{N}}, {t}_{N})\}$, we use $\mathcal{F}_{b}$ to predict the coordinates $\hat{\vec{r}}_1$ of the bounding box of the first room specified by ${\vec{c}}_1$ and ${t}_1$, where $\hat{\vec{r}}_{1}= \mathcal{F}_{b}(\mat{S}_0, {\vec{c}}_{1},{t}_{1} )$ with $\mat{S}_{0}={\mat{B}}$. Furthermore, we obtain the room mask using $\hat{\mat{M}}_{1}= \mathcal{F}_{m}(\hat{\vec{r}}_{1}) \in \mathbb{R}^{128\times128}$. The output of the generator at the first step is computed by
\begin{equation}
    \hat{\mat{S}}_{1}= {\mat{S}}_{0} \times (\vec{1} - \hat{\mat{M}}_{1} ) + \hat{\mat{M}}_{1} \times t_{1},
\label{blending}
\end{equation}
where $\vec{1}\in \mathbb{R}^{128\times128}$ represents a matrix with all elements being 1. During the prediction in the second step, the generator takes as input $\hat{\mat{S}}_{1}$, ${\vec{c}}_{2}$, ${t}_{2}$ and outputs 
$\hat{\mat{S}}_{2}= \hat{\mat{S}}_{1} \times (\vec{1} - \hat{\mat{M}}_{2} ) + \hat{\mat{M}}_{2} \times t_{2}$ with $ \hat{\mat{M}}_{2}=\mathcal{F}_{m}(\hat{\vec{r}}_{2})=\mathcal{F}_{m}\bigl(\mathcal{F}_{b}(\hat{\mat{S}}_1, {\vec{c}}_{2},{t}_{2} )\bigr)$. The same procedure is iterated (for totally $\hat{N}$ steps) until the entire design is accomplished. 

\textit{\textbf{Discriminator:}} The discriminator retains the similar backbone as $\mathcal{F}_{b}$ in the generator, except that the last two convolution layers are dropped. The discriminator is employed to distinguish whether a sequence of design states comes from the generator or the ground truth. Specifically, a sequence of predicted design states $ \hat{\mat{S}} = \{\hat{\vec{S}}_1, \hat{\vec{S}}_1, \ldots, \hat{\vec{S}}_{\hat{N}}\}$ should be recognized as `FALSE', while a sequence from the ground truth with the same order of room types (\ie,$\{ {t}_1, {t}_2, \ldots, {t}_{\hat{N}}\}$) should be predicted as `TRUE'. 

For training, we define a loss on top of WGAN-GP \cite{gulrajani2017improved} loss, \ie,
\begin{equation}
\begin{aligned}
\mathcal{L} = &\mathbb{E}_{\hat{\mat{S}} \sim \mathbb{P}_g}[D(\hat{\mat{S}})] -\mathbb{E}_{{\mat{S}} \sim \mathbb{P}_r}[D(\mat{S})] \\
&+\lambda_1 \mathbb{E}_{\bar{\mat{S}} \sim \mathbb{P}_{\bar{\mat{S}}}}(||\bigtriangledown_{\bar{\mat{S}} }D(\bar{\mat{S}})||_{2} - 1)^2 + \lambda_2 \mathcal{L}_{s},
\end{aligned}
\label{equ:room:partition:loss}
\end{equation}
where $\mathbb{P}_g$ and $\mathbb{P}_r$ represent the generated and real data distribution, respectively. $\bar{\mat{S}} \sim \mathbb{P}_{\bar{\mat{S}}}$ denotes a random interpolation between $\hat{\mat{S}}$ and ${\mat{S}}$, and is used for a gradient penalty with $\lambda_1=10$. $||\cdot||_2$ is the $\ell_2$-norm. The last term is introduced to explicitly regularize the bounding boxes with $\lambda_2=100$:
\begin{equation}
\mathcal{L}_{s} \!=\! \sum_{j=1}^{N} l_j, \,\,\,\,
l_j \!=\! \!\left\{ 
\!\!\!\!\begin{array}{rcl}
 0.5(\vec{r}_j-\hat{\vec{r}}_j),& \text{if } ||\vec{r}_j-\hat{\vec{r}}_j||_1 < 1 \\
 ||\vec{r}_j-\hat{\vec{r}}_j||_1-0.5,& otherwise
\end{array} \right.
\end{equation} 
where $\vec{r}_j$ is the ground truth and $||\cdot||_1$ is the $\ell_1$-norm.

At the inference stage, our procedural model predicts room areas in a step-wise manner, which hence allows user input at an arbitrary intermediate step, such as modifying room types, changing room centers or even the already predicted room areas. Gaps 
may exist between the predicted rooms. Thus, we employ a simple post-processing step as in \cite{sun2019learning} and detail it in the supplementary material.  

\section{Experiments}
\subsection{Datasets}
We conduct experiments on two commonly used datasets, RPLAN\cite{Wu_DeepLayout_2019} and LIFULL\cite{lifull}. \textbf{RPLAN} is collected from the real-world residential buildings in the Asian real estate market,
which contains over 80k floorplans and 13 types of rooms\footnote{LivingRoom, MasterRoom, Kitchen, Bathroom, DiningRoom, ChildRoom, StudyRoom, SecondRoom, GuestRoom, Balcony, Entrance, Storage, Wall-in.}. All floorplans in RPLAN are axis-aligned and pre-processed to the same scale. The training-validation-test split of the dataset is 70\%–15\%–15\%~\cite{hu2020graph2plan}.
\textbf{LIFULL} HOME’s dataset offers approximately five million apartment floorplans from the Japanese house market. The original dataset is given in the form of images, but a subset has been parsed into vector format by \cite{liu2017raster}. 
We select a subset with $4$--$10$ rooms. This specific dataset consists of approximately 54k floorplans and 9 room types\footnote{LivingRoom, Kitchen, Bedroom, Bathroom, Office, Balcony, Hallway, OtherRoom.}, among which 85\% (randomly sampled) of the data serves as the training set while the remaining for testing.

\subsection{Metrics}
Although each floorplan only contains one design (\ie, one boundary corresponds to one design), our model can predict the distribution of plausible designs, offering design diversity and alternative choices. 
To evaluate the predicted distribution, we employ the \textit{Fr\'{e}chet Inception Distance} (FID) \cite{heusel2017gans} to calculate the distance between two distributions, which was also used in floorplan generation \cite{nauata2020house, nauata2021house}.
Built on FID, we introduce three metrics \textit{$FID_{img}$}, \textit{$FID_{area}$} and \textit{$FID_{type}$}:
(\emph{i}) \textit{$FID_{img}$}: computed on rendered images to evaluate the distributional differences of the generated and true images.
(\emph{ii}) \textit{$FID_{area}$}: to evaluate the distributional differences of room areas. Each layout is represented by a $1 \times K$ vector $area_{i}$, with its $k$-th element $area_{i,k}$ representing the average area of the $k$-th type of rooms in the $i$-th floorplan. $K$ denotes the number of room types.
(\emph{iii}) \textit{$FID_{type}$}: to calculate the distributional differences of room numbers against room types.
Each layout is represented by a $1 \times K$ vector $type_{i}$, whose element $type_{i,k}$ represents the number of rooms under the $k$-th type in the $i$-th floorplan.
It is worthwhile to mention that the above three metrics are not biased towards our method as they are not involved in the training process at all.

\subsection{Baselines}
We choose three methods as baselines. Rplan\footnote{We rename the approach RPLAN as Rplan in order to distinguish it from the dataset RPLAN.} \cite{Wu_DeepLayout_2019} is an image-based method, which outputs a floorplan given a boundary. We further use a variant of Rplan, named Rplan$^*$, which also takes as input the room types and centers. HouseGAN++ \cite{nauata2021house} is a graph-constrained approach, which treats rooms as nodes and requires node types and connectivity as input. It generates mask images for all nodes according to the graph, then blends them to form the layout. Graph2plan \cite{hu2020graph2plan} requires a boundary and a room relation graph as input, where the graph includes information on 
room size, room center, room type and their connections. All floorplans are rendered in the same way as \cite{hu2020graph2plan}, including door and window placements.

\begin{table}
  \centering
  \begin{adjustbox}{width=\columnwidth,center}
  \begin{tabular}{ p{1.1cm} | p{1.9cm} p{1.1cm} p{1.7cm} p{1.9cm}}
    \toprule
    Dataset & Method  & $FID_{img}$ & $FID_{area}$ &  $FID_{type}$ \\
    \midrule
     \multirow{7}{*} {RPLAN} &  HouseGAN++ & 51.33 & 1.36 $\times 10^8$ & 0.038  \\
     &  Rplan & 4.1 & 2.29 $\times 10^5$ & 0.58  \\ 
     &  Our$_{III}$ & 1.22 & 3.13 $\times 10^4$ & 0.05 \\
     &  Our$_{II}$ & 0.72 & 1.09 $\times 10^4$  & 0.03 \\
     &  Rplan* & \bf{0.11} & 8.62 $\times 10^{3}$ & 6.4 $\times 10^{-3}$  \\
     &  Graph2Plan & 0.62 & 8.82 $\times 10^{3}$ & 2.70 $\times 10^{-4}$  \\
     &  Our$_{I}$ & 0.16 & \bm { $4.89 \times 10^{2}$} & \bm {$4.44 \times 10^{-6}$} \\
    \midrule
     \multirow{6}{*} {LIFULL}
     &  Rplan & 50.19 & 4.29 $\times 10^{6}$ & 5.15 \\
     &  Our$_{III}$ & 37.35 & 9.81 $\times 10^{5}$  & 2.52 \\
     &  Our$_{II}$ & 32.65 & $7.75 \times 10^{5}$  & 2.14 \\
     &  Rplan* & 1.43 & 5.62 $\times 10^{5}$ & 0.064 \\
     &  Graph2Plan & 0.64 & 2.87 $\times 10^{3}$ & $2.63 \times 10^{-5}$ \\
     &  Our$_{I}$ & \bm{$0.38$} & \bm { $2.07 \times 10^{3}$} & 
     \bm {$3.59 \times 10^{-6}$} \\

    \bottomrule
  \end{tabular}
\end{adjustbox}
  \caption{FID-based metrics on RPLAN and LIFULL.}
  \label{tab:quantitative_result}
  \vspace{-0.4cm}
\end{table}

\subsection{Quantitative Evaluations}
\textbf{Ablation Study}. iPLAN is capable of fully automated generation while allowing for user interactions at different stages. Based on the amount of information received from the designer, three variants are evaluated: (v1) \emph{Our$_{I}$} takes as input the boundary, room types, and room centers; (v2) \emph{Our$_{II}$} takes as input the boundary and room types; (v3) \emph{Our$_{III}$} takes as input only the boundary. The variants show different levels of automation/user interactivity, demonstrating iPLAN's flexibility when the depth of human involvement varies, from little human input (\eg, iPLAN freely explores designs), to step-to-step guidance (\eg, where to put each room). The implementation details are in the supplementary document. \Cref{tab:quantitative_result} shows the results. iPLAN achieves good results on all three settings, among which Our$_{I}$ achieves the best results and  Our$_{II}$ is better than Our$_{III}$. 
This is natural as more prior information helps for better prediction. 
Considering RPLAN and LIFULL are significantly different in terms of the overall shapes of boundaries and rooms, iPLAN can indeed handle different data distributions well.

\begin{figure*}[tb]
  \centering
  \includegraphics[width=0.97\textwidth]{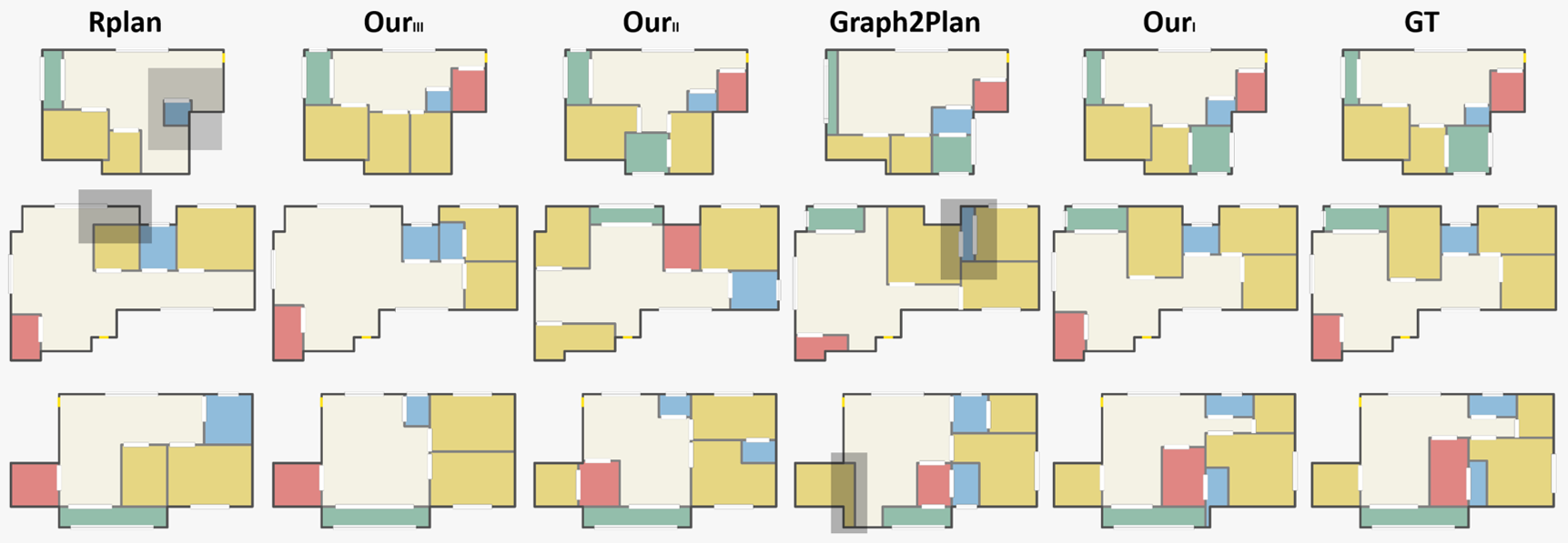}
  \caption{Qualitative comparisons. Shaded areas indicate design choices that are questionable and  unobserved in data.}
  \label{fig:qualitative_eva_rplan}
  \vspace{-0.4cm}
\end{figure*}

\textbf{Comparisons on RPLAN}
iPLAN in general outperforms all baseline methods by large margins shown in \Cref{tab:quantitative_result}. Looking closely, HouseGAN++ achieves worse results, except $FID_{type}$ with a small margin, than other methods. The reason is that its layouts are not bounded by boundaries. By taking the boundary as input, both Rplan and Our$_{III}$ predict room centers and regions successively, but Our$_{III}$ outperforms Rplan on all three metrics.  Furthermore, when both the boundary and room types are given, Our$_{II}$ outperforms the first three methods. Finally, Rplan$^*$, Graph2Plan and Our$_{I}$ are given the full information (boundary, room types and centers). Our$_{I}$ further improves on all metrics, where Rplan$^*$ achieves a slightly better score on $FID_{img}$. We observe that the step-wise prediction in iPLAN is less likely to generate partitioning ambiguity (allocating a space to multiple rooms), while is observed in Rplan$^*$. Also, learning conditional probabilities (as iPLAN does) as oppose to a joint probability of multiple factors from final designs (as Rplan$^*$ and Graph2Plan do) is easier, where the decomposition of final designs can be seen as a data augmentation strategy.

\textbf{Comparisons on LIFULL}
Similar comparisons have been done on LIFULL (\Cref{tab:quantitative_result}). We skip HouseGAN++ because it needs door locations in the input but there is no such data in LIFULL. Compared with Rplan$^*$, Our$_{I}$ and Graph2Plan, room centers are not provided as priors for Rplan, Our$_{III}$ and Our$_{II}$, leading to a large performance gap between these two groups. It is understandable because LIFULL is more challenging/heterogeneous with multi-scale samples and nested rooms. Next, Our$_{II}$ outperforms Our$_{III}$ on all three metrics because more prior knowledge (\ie, room types) is given, and both are better than Rplan. Furthermore, when the room centers are provided, inaccurate wall prediction affects Rplan$^*$. Different from the results on RPLAN, Graph2Plan outperforms Rplan$^*$ on LIFULL. This is because Graph2Plan takes spacial room relations as input which is more robust for such a multi-scale dataset. Our$_{I}$ is still better than Rplan$^*$ and Graph2Plan.
Therefore, we can conclude that, when the same amount of prior knowledge is given, iPLAN achieves the best performance.

\subsection{Qualitative Evaluation}

To show intuitive results, qualitative comparisons are given in \Cref{fig:qualitative_eva_rplan}. Each color represents one room type (see color indication in \cref{fig:auto_interactive_generation}). The shaded boxes indicate questionable design areas. Specifically, in the first row, Rplan designs an isolated bathroom at a corner in the first image, which is a strange design and not observed in the data. In the second row, the first image by Rplan allocates an unreasonable space (shaded) to the public area, which however should be assigned to the neighboring bedroom. Similarly, the layout produced by Graph2Plan consists of 3 bedrooms (second row), but the bathroom is embedded among them (shaded), which means people need to pass through one bedroom to go to the toilet. Besides, the bathroom is too small. As for the last row, the room area is not properly predicted by Graph2plan because the region (shaded) at the bottom-left corner is too narrow.

By comparing Our$_I$, Our$_{II}$ and Our$_{III}$, it is apparent that Our$_I$ is closest to GT as a result of the strong input constraints. In contrast, when less constraints are imposed, the diversity is significant, as depicted in Our$_{II}$ and Our$_{III}$.
More results and analysis are shown in the supplementary material.

\subsection{User Interactions}
\begin{figure*}[tb]
  \centering
  \includegraphics[width=0.98\textwidth]{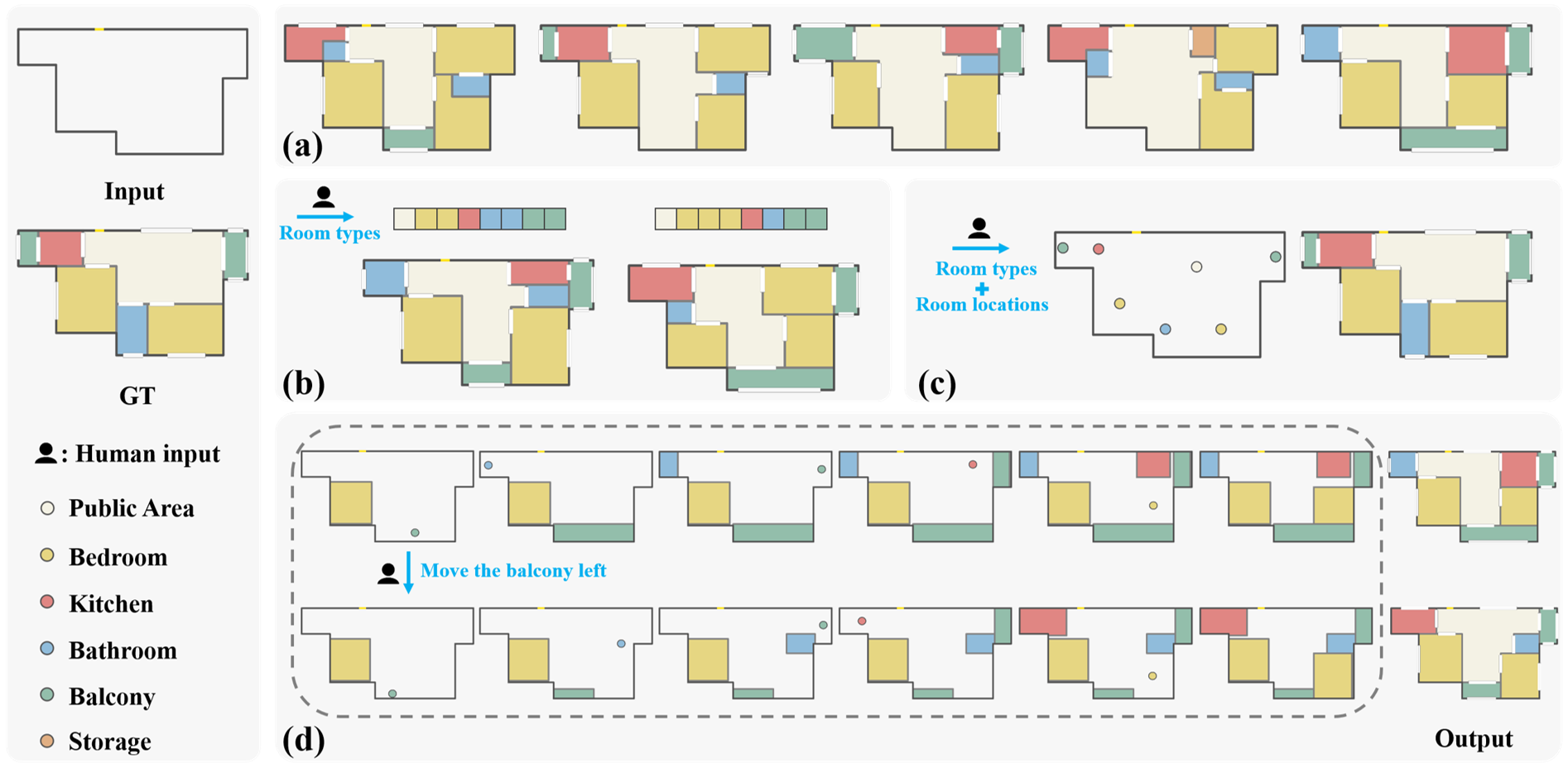}
  \caption{User interactions. (a)--(c) indicate interactions via providing different levels of human input; (d) shows fine-grained interaction in a step-wise generation.}
  \label{fig:auto_interactive_generation}
  \vspace{-0.4cm}
\end{figure*}

The performance of iPLAN under different levels of human interactions is showcased in~\Cref{fig:auto_interactive_generation}. \Cref{fig:auto_interactive_generation}(\emph{a}) is fully automated generation without human input other than a boundary. iPLAN can generate diverse layouts,  
with a varying number of rooms and different room types/areas. \Cref{fig:auto_interactive_generation}(\emph{b}) shows the user can decide the room types, where
the final layouts are different when the order of room types varies, showing the flexibility of iPLAN. We further provide iPLAN with more priors in Fig.~\ref{fig:auto_interactive_generation}(\emph{c}), \ie, the boundary, room types and centers, leading to a planned layout that is nearly the same as the ground truth. We also evaluate iPLAN by introducing a user input at an intermediate step (Fig.~\ref{fig:auto_interactive_generation}(\emph{d})), where the balcony is moved to the left and consequently a different layout is obtained. \Cref{fig:auto_interactive_generation} demonstrates the interactive ability of iPLAN, which is crucial in settings where the designer leads the design.

\begin{figure}[tb]
  \centering
  \includegraphics[width=0.49\textwidth]{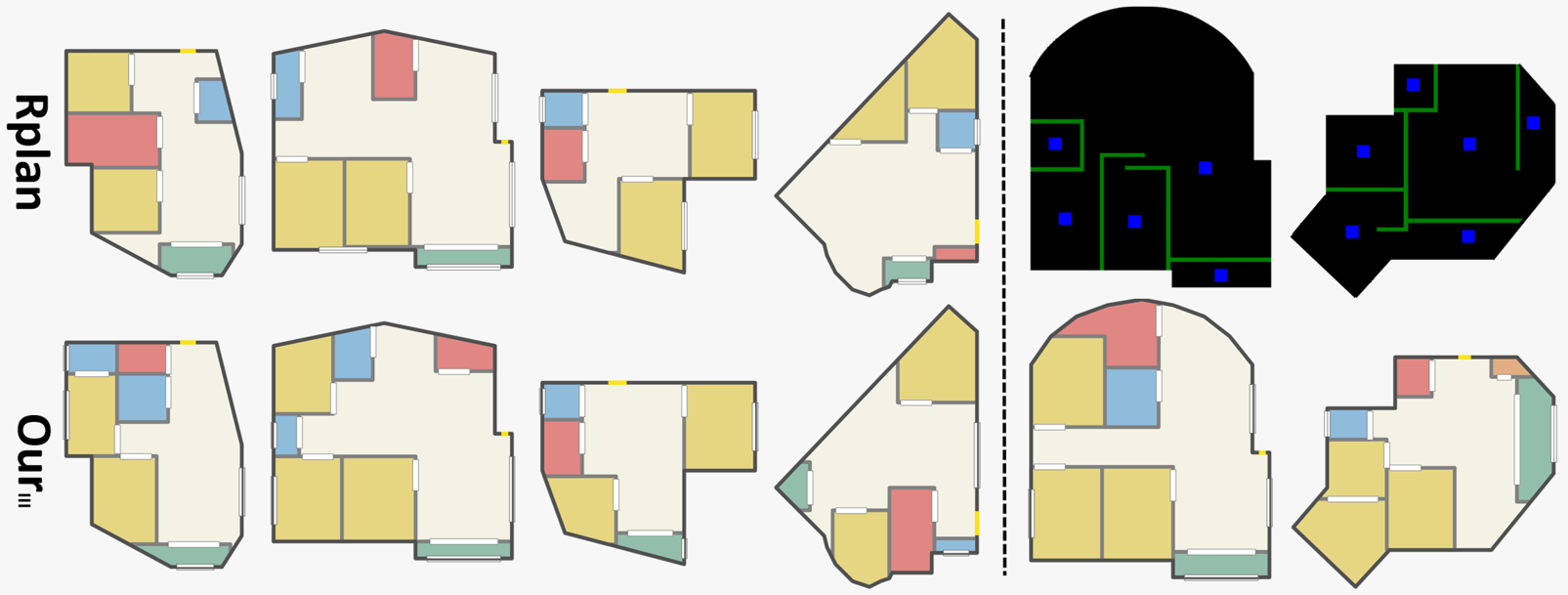}
  \caption{Layouts with non-axis aligned edges. Left: successful examples. Right: Rplan fails but our model succeeds.}
  \label{fig:irregular}
  \vspace{-0.4cm}
\end{figure}

\subsection{Generalizability}

\begin{table}[tb]
  \centering
  \begin{adjustbox}{width=0.91\columnwidth,center}
  \begin{tabular}{ p{1.1cm}|p{1.1cm} p{1.1cm} p{1.7cm} p{1.9cm}}
    \toprule
    Dataset & Method  & $FID_{img}$ & $FID_{area}$ &  $FID_{type}$ \\
    \midrule
     \multirow{4}{*} {RPLAN}
     & Our$_{II}^{8}$ & 1.2 & $1.36  \times 10^{4}$  & 0.06 \\  
     & Our$_{I}^{8}$ & 0.2 &  $ 2.69 \times 10^{2}$ & 
     $1.25 \times 10^{-5}$ \\
     \cline{2-5}
     & Our$_{II}^{7\sim 8}$ & 2.28 &  $4.87 \times 10^{4}$ & 
     0.18 \\
     & Our$_{I}^{7\sim 8}$ & 0.24 &  $6.89 \times 10^{2}$ & 
     $1.54 \times 10^{-5}$ \\
    \bottomrule
  \end{tabular}
  \end{adjustbox}
  \caption{Generalization results on RPLAN.}
  \label{tab:generalization_results_rplan}
  \vspace{-0.4cm}
\end{table}

\textbf{Quantitative results} To further push iPLAN,
we quantitatively evaluate the generalizability of iPLAN by setting up two
groups of experiments on RPLAN. In the first group, we select 51322 layouts consisting of $4\sim 7$ rooms for training and randomly choose 12,000 8-room layouts for testing. In the second group, we consider a more challenging task by randomly selecting 26574 layouts containing $ 4 \sim 6$ rooms for training and 12000 layouts containing $ 7 \sim 8 $ rooms for testing. Note that in both experiments, the layouts for test include more rooms than the ones used for training. 

Quantitative results are reported in \Cref{tab:generalization_results_rplan}. Again, we can observe that Our$_{I}$ outperforms Our$_{II}$,
which is consistent with our intuition since more prior knowledge is provided to Our$_{I}$. Further, the results in first group are better than those in the second group, which is reasonable because the generalization setting in the second group is more challenging.
However, both groups achieve comparable performance to \Cref{tab:quantitative_result} and still outperform the baselines.

\textbf{Qualitative Results}
An interesting test is to see if iPLAN trained on RPLAN can be generalized to unseen types of boundaries. The qualitative results are also provided to verify the generalizability of iPLAN. Note that RPLAN only contains layouts with straight axis-aligned edges in the boundaries, so we consider the boundaries with non-axis aligned edges and curves.
Since HouseGAN++ cannot design a layout for a specific building boundary and Graph2Plan is restricted to boundaries with axis-aligned edges, 
we only compare iPLAN with Rplan.

Figure \ref{fig:irregular} shows several examples of non-axis aligned design produced by Rplan and iPLAN. The left lists four successful cases on both approaches and the right shows two examples where Rplan fails but iPLAN succeeds. Even when successful, Rplan often predicts isolated rooms which divide the public areas into strange shapes and reduce their usability (the first two examples), while iPLAN utilizes the space better. As for the failures of Rplan, we give two examples of room center estimation and wall prediction, where it fails to reasonably fill and partition the space, a problem from which iPLAN does not suffer. Please refer to the supplementary document for more examples.

\section{Limitation \& Conclusion}
While iPLAN is able to handle some irregular boundaries, it cannot cope with extremely irregular ones. One important future direction would be to handle complex environments, e.g., non-axis aligned boundaries and compound and nested rooms, which is crucial to generalize iPLAN to public spaces such as train stations and shopping malls.

In this paper, we proposed a novel human-in-the-loop generative model iPLAN to learn professional designs in a stage-to-stage fashion while respecting design principles. While being capable of fully automated generation, iPLAN allows close interactions with humans by accepting user guidance at every stage and automatically suggesting possible designs accordingly. Comprehensive evaluations on two benchmark datasets RPLAN and LIFULL show that iPLAN outperforms the state-of-the-art methods, both quantitatively and qualitatively. Importantly, iPLAN has exhibited strong generalization capability to unseen design tasks and boundary inputs.

\noindent
\textbf{Acknowledgements:} In this paper, we used "LIFULL HOME'S Snapshot Data of Rentals" provided by LIFULL Co., Ltd. via IDR Dataset Service of National Institute of Informatics. We thank Jing Li for her input on the design practice. This   project   has   received   funding   from   the   European
Union’s Horizon 2020 research and innovation programme
under grant agreement No 899739 CrowdDNA and the Marie Skłodowska-Curie grant
agreement No 101018395. Feixiang He has been supported by UKRI PhD studentship [EP/R513258/1, 2218576].

{\small
\bibliographystyle{ieee_fullname}
\bibliography{egbib}

\begin{thebibliography}{10}\itemsep=-1pt

\bibitem{ashual2019specifying}
Oron Ashual and Lior Wolf.
\newblock Specifying object attributes and relations in interactive scene
  generation.
\newblock In {\em Proceedings of the IEEE/CVF International Conference on
  Computer Vision}, pages 4561--4569, 2019.

\bibitem{bao2013generating}
Fan Bao, Dong-Ming Yan, Niloy~J Mitra, and Peter Wonka.
\newblock Generating and exploring good building layouts.
\newblock {\em ACM Transactions on Graphics (TOG)}, 32(4):1--10, 2013.

\bibitem{brock2018large}
Andrew Brock, Jeff Donahue, and Karen Simonyan.
\newblock Large scale gan training for high fidelity natural image synthesis.
\newblock {\em arXiv preprint arXiv:1809.11096}, 2018.

\bibitem{chaillou2020archigan}
Stanislas Chaillou.
\newblock Archigan: Artificial intelligence x architecture.
\newblock In {\em Architectural intelligence}, pages 117--127. Springer, 2020.

\bibitem{chen2017deeplab}
Liang-Chieh Chen, George Papandreou, Iasonas Kokkinos, Kevin Murphy, and Alan~L
  Yuille.
\newblock Deeplab: Semantic image segmentation with deep convolutional nets,
  atrous convolution, and fully connected crfs.
\newblock {\em IEEE transactions on pattern analysis and machine intelligence},
  40(4):834--848, 2017.

\bibitem{fisher2012example}
Matthew Fisher, Daniel Ritchie, Manolis Savva, Thomas Funkhouser, and Pat
  Hanrahan.
\newblock Example-based synthesis of 3d object arrangements.
\newblock {\em ACM Transactions on Graphics (TOG)}, 31(6):1--11, 2012.

\bibitem{goodfellow2014generative}
Ian Goodfellow, Jean Pouget-Abadie, Mehdi Mirza, Bing Xu, David Warde-Farley,
  Sherjil Ozair, Aaron Courville, and Yoshua Bengio.
\newblock Generative adversarial nets.
\newblock {\em Advances in neural information processing systems}, 27, 2014.

\bibitem{gulrajani2017improved}
Ishaan Gulrajani, Faruk Ahmed, Martin Arjovsky, Vincent Dumoulin, and Aaron
  Courville.
\newblock Improved training of wasserstein gans.
\newblock {\em arXiv preprint arXiv:1704.00028}, 2017.

\bibitem{harada1995interactive}
Mikako Harada, Andrew Witkin, and David Baraff.
\newblock Interactive physically-based manipulation of discrete/continuous
  models.
\newblock In {\em Proceedings of the 22nd annual conference on Computer
  graphics and interactive techniques}, pages 199--208, 1995.

\bibitem{he2016deep}
Kaiming He, Xiangyu Zhang, Shaoqing Ren, and Jian Sun.
\newblock Deep residual learning for image recognition.
\newblock In {\em Proceedings of the IEEE conference on computer vision and
  pattern recognition}, pages 770--778, 2016.

\bibitem{heusel2017gans}
Martin Heusel, Hubert Ramsauer, Thomas Unterthiner, Bernhard Nessler, and Sepp
  Hochreiter.
\newblock Gans trained by a two time-scale update rule converge to a local nash
  equilibrium.
\newblock {\em Advances in neural information processing systems}, 30, 2017.

\bibitem{hu2020graph2plan}
Ruizhen Hu, Zeyu Huang, Yuhan Tang, Oliver Van~Kaick, Hao Zhang, and Hui Huang.
\newblock Graph2plan: Learning floorplan generation from layout graphs.
\newblock {\em ACM Transactions on Graphics (TOG)}, 39(4):118--1, 2020.

\bibitem{huang2018multimodal}
Xun Huang, Ming-Yu Liu, Serge Belongie, and Jan Kautz.
\newblock Multimodal unsupervised image-to-image translation.
\newblock In {\em Proceedings of the European conference on computer vision
  (ECCV)}, pages 172--189, 2018.

\bibitem{isola2017image}
Phillip Isola, Jun-Yan Zhu, Tinghui Zhou, and Alexei~A Efros.
\newblock Image-to-image translation with conditional adversarial networks.
\newblock In {\em Proceedings of the IEEE conference on computer vision and
  pattern recognition}, pages 1125--1134, 2017.

\bibitem{johnson2018image}
Justin Johnson, Agrim Gupta, and Li Fei-Fei.
\newblock Image generation from scene graphs.
\newblock In {\em Proceedings of the IEEE conference on computer vision and
  pattern recognition}, pages 1219--1228, 2018.

\bibitem{karras2017progressive}
Tero Karras, Timo Aila, Samuli Laine, and Jaakko Lehtinen.
\newblock Progressive growing of gans for improved quality, stability, and
  variation.
\newblock {\em arXiv preprint arXiv:1710.10196}, 2017.

\bibitem{karras2020training}
Tero Karras, Miika Aittala, Janne Hellsten, Samuli Laine, Jaakko Lehtinen, and
  Timo Aila.
\newblock Training generative adversarial networks with limited data.
\newblock {\em arXiv preprint arXiv:2006.06676}, 2020.

\bibitem{karras2019style}
Tero Karras, Samuli Laine, and Timo Aila.
\newblock A style-based generator architecture for generative adversarial
  networks.
\newblock In {\em Proceedings of the IEEE/CVF Conference on Computer Vision and
  Pattern Recognition}, pages 4401--4410, 2019.

\bibitem{karras2020analyzing}
Tero Karras, Samuli Laine, Miika Aittala, Janne Hellsten, Jaakko Lehtinen, and
  Timo Aila.
\newblock Analyzing and improving the image quality of stylegan.
\newblock In {\em Proceedings of the IEEE/CVF Conference on Computer Vision and
  Pattern Recognition}, pages 8110--8119, 2020.

\bibitem{kingma2013auto}
Diederik~P Kingma and Max Welling.
\newblock Auto-encoding variational bayes.
\newblock {\em arXiv preprint arXiv:1312.6114}, 2013.

\bibitem{li2019pastegan}
Yikang Li, Tao Ma, Yeqi Bai, Nan Duan, Sining Wei, and Xiaogang Wang.
\newblock Pastegan: A semi-parametric method to generate image from scene
  graph.
\newblock {\em Advances in Neural Information Processing Systems},
  32:3948--3958, 2019.

\bibitem{lifull}
Ltd. LIFULL~Co.
\newblock Lifull home's snapshot data of rentals, nov 2015.

\bibitem{liu2017raster}
Chen Liu, Jiajun Wu, Pushmeet Kohli, and Yasutaka Furukawa.
\newblock Raster-to-vector: Revisiting floorplan transformation.
\newblock In {\em Proceedings of the IEEE International Conference on Computer
  Vision}, pages 2195--2203, 2017.

\bibitem{merrell2011interactive}
Paul Merrell, Eric Schkufza, Zeyang Li, Maneesh Agrawala, and Vladlen Koltun.
\newblock Interactive furniture layout using interior design guidelines.
\newblock {\em ACM transactions on graphics (TOG)}, 30(4):1--10, 2011.

\bibitem{muller2006procedural}
Pascal M{\"u}ller, Peter Wonka, Simon Haegler, Andreas Ulmer, and Luc Van~Gool.
\newblock Procedural modeling of buildings.
\newblock In {\em ACM SIGGRAPH 2006 Papers}, pages 614--623. 2006.

\bibitem{nauata2020house}
Nelson Nauata, Kai-Hung Chang, Chin-Yi Cheng, Greg Mori, and Yasutaka Furukawa.
\newblock House-gan: Relational generative adversarial networks for
  graph-constrained house layout generation.
\newblock In {\em European Conference on Computer Vision}, pages 162--177.
  Springer, 2020.

\bibitem{nauata2021house}
Nelson Nauata, Sepidehsadat Hosseini, Kai-Hung Chang, Hang Chu, Chin-Yi Cheng,
  and Yasutaka Furukawa.
\newblock House-gan++: Generative adversarial layout refinement networks.
\newblock {\em arXiv preprint arXiv:2103.02574}, 2021.

\bibitem{peng2014computing}
Chi-Han Peng, Yong-Liang Yang, and Peter Wonka.
\newblock Computing layouts with deformable templates.
\newblock {\em ACM Transactions on Graphics (TOG)}, 33(4):1--11, 2014.

\bibitem{rengel2011interior}
Roberto~J Rengel.
\newblock {\em The interior plan: Concepts and exercises}.
\newblock A\&C Black, 2011.

\bibitem{richardson2021encoding}
Elad Richardson, Yuval Alaluf, Or Patashnik, Yotam Nitzan, Yaniv Azar, Stav
  Shapiro, and Daniel Cohen-Or.
\newblock Encoding in style: a stylegan encoder for image-to-image translation.
\newblock In {\em Proceedings of the IEEE/CVF Conference on Computer Vision and
  Pattern Recognition}, pages 2287--2296, 2021.

\bibitem{ritchie2019fast}
Daniel Ritchie, Kai Wang, and Yu-an Lin.
\newblock Fast and flexible indoor scene synthesis via deep convolutional
  generative models.
\newblock In {\em Proceedings of the IEEE/CVF Conference on Computer Vision and
  Pattern Recognition}, pages 6182--6190, 2019.

\bibitem{sohn2015learning}
Kihyuk Sohn, Honglak Lee, and Xinchen Yan.
\newblock Learning structured output representation using deep conditional
  generative models.
\newblock {\em Advances in neural information processing systems},
  28:3483--3491, 2015.

\bibitem{sun2019learning}
Chun-Yu Sun, Qian-Fang Zou, Xin Tong, and Yang Liu.
\newblock Learning adaptive hierarchical cuboid abstractions of 3d shape
  collections.
\newblock {\em ACM Transactions on Graphics (TOG)}, 38(6):1--13, 2019.

\bibitem{wang2019planit}
Kai Wang, Yu-An Lin, Ben Weissmann, Manolis Savva, Angel~X Chang, and Daniel
  Ritchie.
\newblock Planit: Planning and instantiating indoor scenes with relation graph
  and spatial prior networks.
\newblock {\em ACM Transactions on Graphics (TOG)}, 38(4):1--15, 2019.

\bibitem{wang2018deep}
Kai Wang, Manolis Savva, Angel~X Chang, and Daniel Ritchie.
\newblock Deep convolutional priors for indoor scene synthesis.
\newblock {\em ACM Transactions on Graphics (TOG)}, 37(4):1--14, 2018.

\bibitem{Wu_DeepLayout_2019}
Wenming Wu, Xiao-Ming Fu, Rui Tang, Yuhan Wang, Yu-Hao Qi, and Ligang Liu.
\newblock Data-driven interior plan generation for residential buildings.
\newblock {\em ACM Transactions on Graphics (SIGGRAPH Asia)}, 38(6), 2019.

\bibitem{yu2011make}
Lap~Fai Yu, Sai~Kit Yeung, Chi~Keung Tang, Demetri Terzopoulos, Tony~F Chan,
  and Stanley~J Osher.
\newblock Make it home: automatic optimization of furniture arrangement.
\newblock {\em ACM Transactions on Graphics (TOG)-Proceedings of ACM SIGGRAPH
  2011, v. 30,(4), July 2011, article no. 86}, 30(4), 2011.

\bibitem{zhang2019self}
Han Zhang, Ian Goodfellow, Dimitris Metaxas, and Augustus Odena.
\newblock Self-attention generative adversarial networks.
\newblock In {\em International conference on machine learning}, pages
  7354--7363. PMLR, 2019.

\bibitem{zhao2016relationship}
Xi Zhao, Ruizhen Hu, Paul Guerrero, Niloy Mitra, and Taku Komura.
\newblock Relationship templates for creating scene variations.
\newblock {\em ACM Transactions on Graphics (TOG)}, 35(6):1--13, 2016.

\bibitem{Zhao2014indexing}
Xi Zhao, He Wang, and Taku Komura.
\newblock Indexing 3d scenes using the interaction bisector surface.
\newblock {\em ACM Trans. Graph.}, 33(3):22:1--22:14, June 2014.

\bibitem{zhu2017unpaired}
Jun-Yan Zhu, Taesung Park, Phillip Isola, and Alexei~A Efros.
\newblock Unpaired image-to-image translation using cycle-consistent
  adversarial networks.
\newblock In {\em Proceedings of the IEEE international conference on computer
  vision}, pages 2223--2232, 2017.

\bibitem{zhu2017multimodal}
Jun-Yan Zhu, Richard Zhang, Deepak Pathak, Trevor Darrell, Alexei~A Efros,
  Oliver Wang, and Eli Shechtman.
\newblock Multimodal image-to-image translation by enforcing bi-cycle
  consistency.
\newblock In {\em Advances in neural information processing systems}, pages
  465--476, 2017.

\bibitem{zhu2020sean}
Peihao Zhu, Rameen Abdal, Yipeng Qin, and Peter Wonka.
\newblock Sean: Image synthesis with semantic region-adaptive normalization.
\newblock In {\em Proceedings of the IEEE/CVF Conference on Computer Vision and
  Pattern Recognition}, pages 5104--5113, 2020.

\end{thebibliography}
}

\clearpage
\appendix

\section{Architecture of BCVAE}
The detailed architecture of BCVAE is illustrated in \cref{tab:BCVAE_architectural_specification}. For a given layout, $\bm{Q}=\{q_{k}\}_{k=1}^{K}$ represents room types, where $K$ denotes the number of room types in $\mat{D}$ and $q_{k} \in \mathbb{Z}$ corresponds to the number of rooms under the $k$-th type. 

Before feeding $\vec{Q}$ into BCVAE, 
a reformulation is implemented. We first determine the largest number of rooms for each type $k$ across the whole dataset and denote it as $q_{k}^* \in \mathbb{Z}$. Then, for each $q_{k} \in \vec{Q}$ ($q_{k}\leq q_{k}^*$), we transform it into a $q_{k}^*$-D vector, i.e., $v_{k}$, whose first $q_{k}$ elements are set as 1 while the remaining elements as 0. By concatenating all transformed vectors, we can obtain an alternative representation of $\vec{Q}$, i.e., $\bm{V}=[\vec{v}_{1}^T \vec{v}_{2}^T \ldots \vec{v}_{K}^T]^{T}$. 

We denote the output of BCVAE as $\bm{\hat{V}}$ and use binary cross entropy as the reconstruction loss:
\begin{equation}
\mathcal{L}_{rec} = \sum_{j=1}^{n\_c} l_j,
l_j = -[v_jlog \hat{v}_j + (1 - v_j)log(1-\hat{v}_j)],
\end{equation} 
where $n\_c=\sum_{i=1}^{K} q_k^*$ represents the length of $\vec{V}$.

The total loss of BCVAE is:
\begin{equation}
    \mathcal{L} = \mathcal{L}_{rec} + \lambda D_{KL}(\mathcal{N}(\vec{\mu},\vec{\Sigma}) || \mathcal{N}(\vec{0},\vec{I})),
\end{equation}
where $D_{KL}$ denotes the Kullback–Leibler (KL) divergence, $\lambda=0.5$.

\begin{table*}[t]
  \centering
  \begin{tabular}{ c c c c }
    \toprule
    Architecture & Layer  & Specification & Output Size \\
    \midrule
    \midrule
     \multirow{6}{*} {embedding network}
     &  $conv\_bn\_relu_1$  & $1\times 16\times 4\times 4 (s=2, p=1)$ & $64\times64\times16$ \\ 
     \cline{2-4}
     &  $conv\_bn\_relu_2$  & $16\times 16\times 4\times 4 (s=2, p=1)$ & $32\times32\times16$ \\ 
     \cline{2-4}
     &  $conv\_bn\_relu_3$  & $16\times 32\times 4\times 4 (s=2, p=1)$ & $16\times16\times32$ \\
     \cline{2-4}
     &  $conv\_bn\_relu_4$ & $32\times 32\times 4\times 4 (s=2, p=1)$ & $8\times8\times32$ \\ 
     \cline{2-4}
     &  $conv\_bn\_relu_5$  & $32\times 16\times 4\times 4 (s=2, p=1)$ & $4\times4\times16$ \\  
     \cline{2-4}
     & $conv\_bn\_relu_6$  & $16\times 16\times 4\times 4 (s=2, p=1)$ & $2\times2\times16$ \\  
     \cline{2-4}
     & $flatten$  & $N/A$ & $1\times64$ \\ 
     \midrule
     \multirow{5}{*} {encoder}
     &  $concat$  & $N/A$ & $1 \times(n\_c + 64)$ \\
     \cline{2-4}
     &  $linear\_relu1$  & $(n\_c+64)\times 128$ & $1 \times128$ \\
     \cline{2-4}
     &  $linear\_relu2$  & $128\times 64$ & $1 \times 64$ \\
     \cline{2-4}
     &  $linear_{31}$  & $64\times32$ & $1 \times32$ \\
     \cline{2-4}
     &  $linear_{32}$  & $64\times 32$ & $1 \times 32$ \\
    \midrule
    \multirow{5}{*} {decoder}
     &  $concat$  & $N/A$ & $1 \times96$ \\
     \cline{2-4}
     &  $linear\_relu1$  & $96 \times96$ & $1 \times96$ \\
     \cline{2-4}
     &  $linear\_relu2$  & $96\times64$ & $1 \times64$ \\
     \cline{2-4}
     &  $linear_3$  & $64\times n\_c$ & $1 \times n\_c$ \\
     \cline{2-4}
     &  $sigmoid$  & $N/A$ & $1 \times n\_c$ \\
    \bottomrule
  \end{tabular}

  \caption{The BCVAE architectural specification. $s$ and $p$ respectively denote stride and padding. $n\_c$ is the dimension of house type. Convolution kernels and layer output are separately specified by $(N_{in} \times N_{out}\times W\times H)$ and $(W\times H \times C)$.}
  \label{tab:BCVAE_architectural_specification}
\end{table*}

\section{Post-Processing for Room Partitioning Prediction}

After the room partitioning prediction, sometimes gaps may exist between the predicted rooms $\mat{\hat{R}} = \{\hat{\vec{r}}_{1}, \hat{\vec{r}}_{2},\ldots, \hat{\vec{r}}_{N} \}$. We employ a simple post-processing step to ensure that the interior area of $\mat{B}$ is fully covered and the room bounding boxes are located within $\mat{B}$. We formulate it as a generic optimization problem:
\begin{equation}
    \mathop{\arg\min}_{\hat{\mat{R}}} \mathcal{L} = \mathop{\arg\min}_{\hat{\mat{R}}} \mathcal{L}_{coverage}(\hat{\mat{R}}, \mat{B}) + \mathcal{L}_{interior}(\hat{\mat{R}}, \mat{B})
    \label{eq:loss_postprocess}
\end{equation}
where $\mathcal{L}_{coverage}$ and $\mathcal{L}_{interior}$ constrain the spatial consistency between $\mat{B}$ and the room bounding box set $\hat{\mat{R}}$ .

To explain $\mathcal{L}_{coverage}$ and $\mathcal{L}_{interior}$ clearly, we introduce a distance function $d(p,\vec{r})$ to measure the coverage of a point $p$ by a box $\vec{r}$:
\begin{equation}
d(p,\vec{r}) \!=\! \!\left\{ 
\!\!\!\!\begin{array}{lcl}
 \quad\quad\quad 0,& \text{if } p \in \Omega_{in}(\vec{r}) \\
 \min_{q \in \Omega_{bd}(\vec{r})}||p -q||,& otherwise
\end{array} \right.
\end{equation} 
where $\Omega_{in}(\vec{r})$ denotes the interior area of the box $\vec{r}$ and  $\Omega_{bd}(\vec{r})$ represents the boundary of $\vec{r}$.

The coverage loss can be defined as:
\begin{equation}
\mathcal{L}_{coverage}(\hat{\mat{R}},\mat{B})= \frac{\sum_{p \in \Omega_{in}(\mat{B})}\min_i  d^2(p,\hat{\vec{r}}_i)}{|\Omega_{in}(\mat{B})|},
\end{equation}
where $|\Omega_{in}|$ is the number of pixels in the set  $\Omega_{in}(\mat{B})$.

The interior loss can be denoted as follows:
\begin{equation}
\mathcal{L}_{interior}(\hat{\mat{R}}, \mat{B})= \frac{\sum_i\sum_{p \in \Omega_{in}(\hat{\vec{r}}_i)} d^2(p,\hat{\mat{B}})}{\sum_i|\Omega_{in}(\hat{\vec{r}}_i)|},
\end{equation}
where $\hat{\mat{B}}$ is the bounding box of the boundary. Note that ${\mat{B}} \subseteq\hat{\mat{B}}$.

Therefore, in the inference stage, we directly adjust the predicted rooms $\mat{\hat{R}} = \{\hat{\vec{r}}_{1}, \hat{\vec{r}}_{2},\ldots, \hat{\vec{r}}_{N} \}$ by minimizing the loss $\mathcal{L}$ in \cref{eq:loss_postprocess}.

\begin{figure*}[tb]
  \centering
  \includegraphics[width=0.92\textwidth]{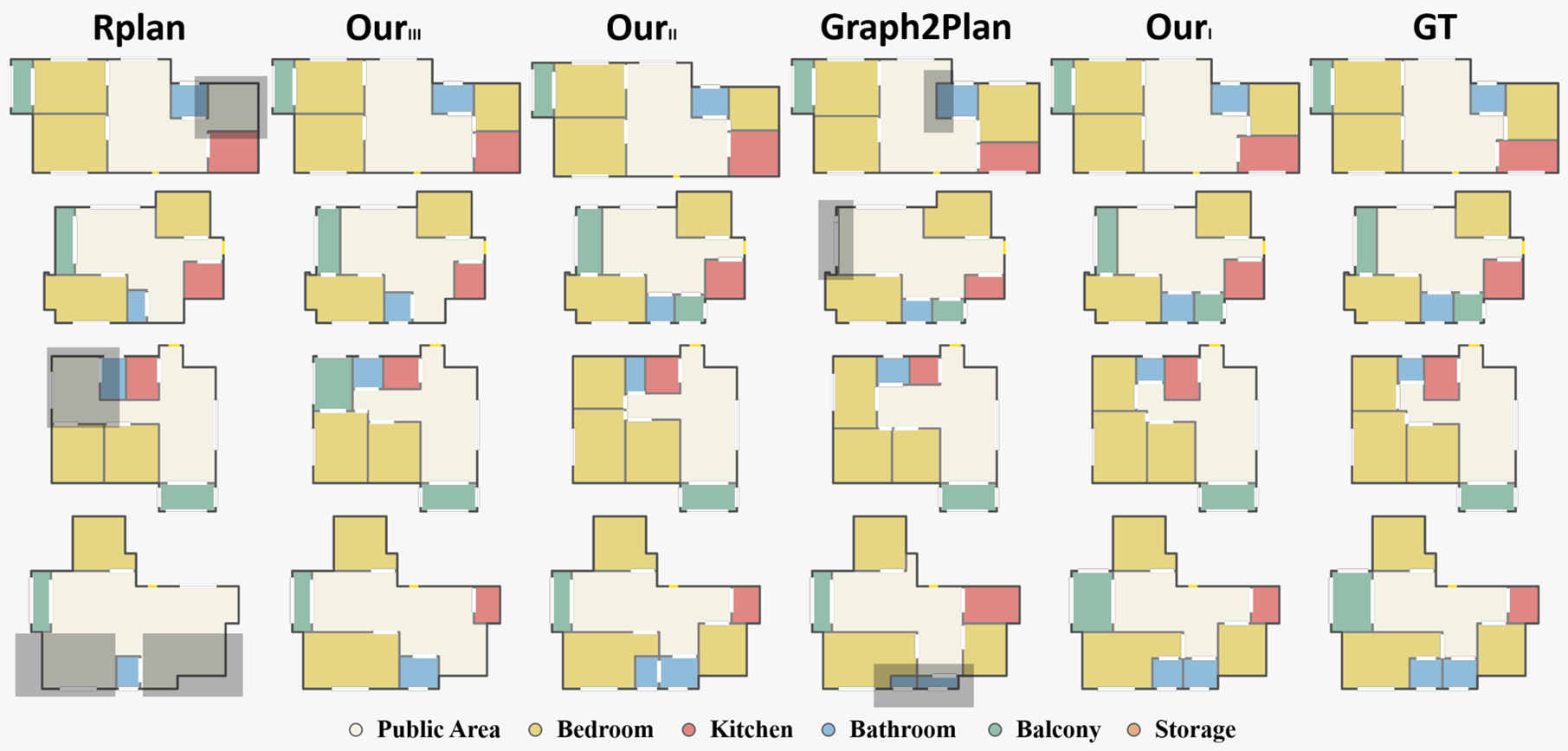}
  \caption{Qualitative comparisons on RPLAN. Shaded areas indicate design choices that are questionable.}
  \label{fig:supp_qualitative_eva_rplan}
\end{figure*}

\begin{figure*}[tb]
  \centering
  \includegraphics[width=0.92\textwidth]{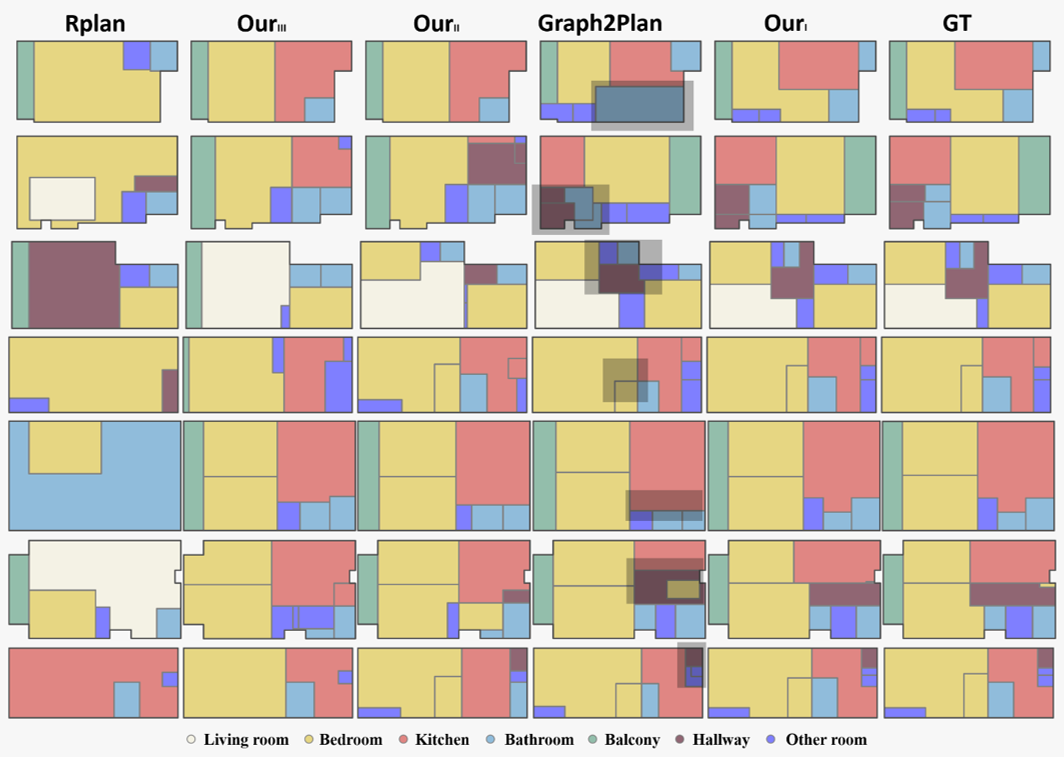}
  \caption{Qualitative comparisons on LIFULL. Shaded areas indicate design choices that are questionable.}
  \label{fig:supp_qualitative_eva_lifull}
\end{figure*}

\begin{figure*}[tb]
  \centering
  \includegraphics[width=0.92\textwidth]{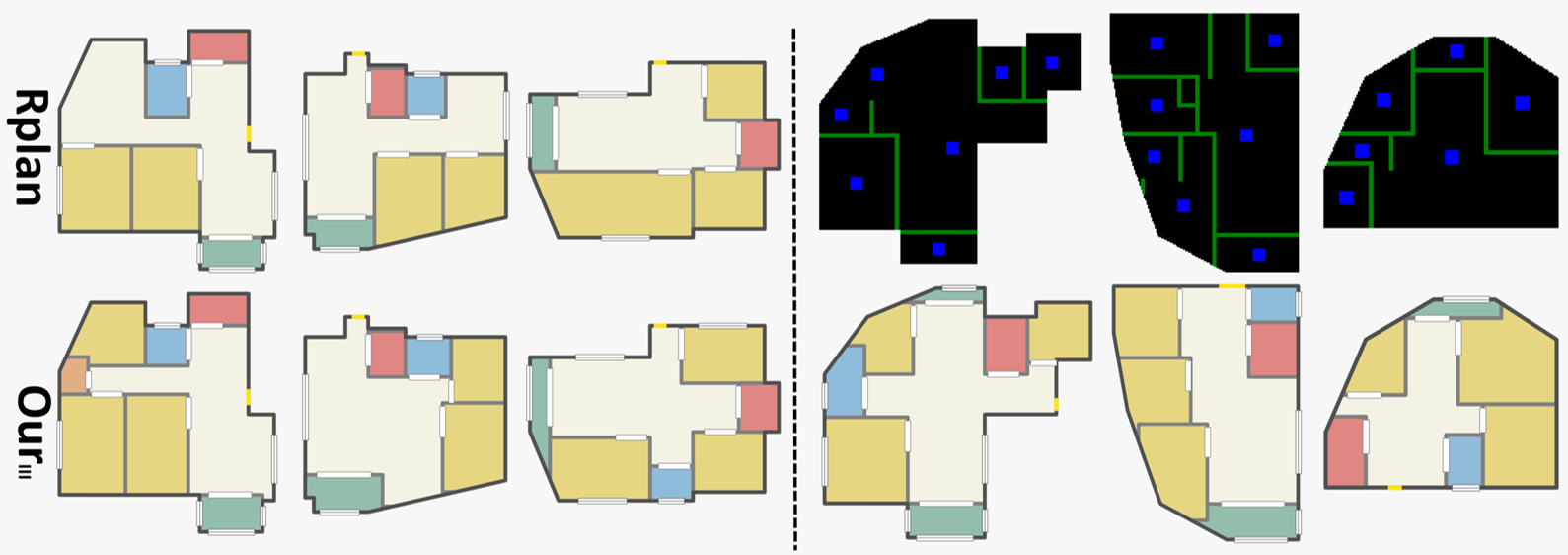}
  \caption{Layouts with non-axis aligned edges.  Left: successful examples. Right: Rplan fails but our model succeeds. }
  \label{fig:supp_irregular}
\end{figure*}

\section{Additional Qualitative Comparisons}
\cref{fig:supp_qualitative_eva_rplan} and \cref{fig:supp_qualitative_eva_lifull} show qualitative results on RPLAN and LIFULL, respectively. In both datasets, Graph2Plan and Our$_I$ are provided with the full human input (including the boundary, room types and room locations), their generated layouts are expected to be similar to the GT. While it is the case for Our$_I$, it doesn't seem to be so for Graph2Plan. The shaded areas on the layouts produced by Graph2Plan show clear differences from the GT layouts. In contrast, the layouts from Our$_I$ are nearly the same as the GT.

In addition, we also compare iPLAN with other methods on a more challenging dataset, LIFULL. 
When only the house boundary is provided, Our$_{III}$ outperforms Rplan. Our$_{II}$ corresponds to the case when the house boundary and the room types are known, which achieves slightly better predictions than Our$_{III}$ as more information is fed.
Furthermore, if the full human input is provided, Our$_{I}$ performs better than Our$_{II}$. Note that Our$_{I}$ is superior to Graph2Plan which is also fed with the full human input.

\section{Additional Generalization Evaluations}
More generalization results on RPLAN are presented in \cref{fig:supp_irregular}. The first row and second row correspond to Rplan and Our$_{III}$, respectively. Our$_{III}$ achieves better results. Consistent with our analysis in the main paper, Rplan is prone to splitting the public area into two main areas, causing potential inconvenience for family activities (the first two columns in \cref{fig:supp_irregular}). Sometimes, Rplan also fails to plan a bathroom in the layout (the third column in \cref{fig:supp_irregular}). Moreover, on some boundaries, Rplan fails to design the layouts (the last three columns in \cref{fig:supp_irregular}). In general, Our$_{III}$ outperforms Rplan when the boundary is non-axis aligned.

\section{Implementation Details}
We have implemented iPLAN in PyTorch. All models are trained and tested on a NVIDIA GeForce RTX 2080 Ti. It takes about two hours to train BCVAE, two days to optimize the room-locating network and one day to train the room area prediction model. 

\end{document}